\pdfoutput=1
\documentclass[11pt]{article}
\usepackage{emnlp2021}
\usepackage{times}
\usepackage{latexsym}

\usepackage[T1]{fontenc}

\usepackage[utf8]{inputenc}

\usepackage{microtype}

\usepackage{amsmath}
\usepackage{comment}
\usepackage{graphicx}
\usepackage[tableposition=top]{caption}
\DeclareCaptionLabelFormat{andtable}{#1~#2  \&  \tablename~\thetable}
\usepackage{subcaption}
\usepackage{float}
\usepackage[ruled,linesnumbered,noend]{algorithm2e}
\usepackage{soul}

\usepackage[textwidth=21mm]{todonotes}
\usepackage{tabularx}

\newcommand{\COMMENT}[1]{}

\title{
The Effect of Efficient Messaging and Input Variability \\ 
on Neural-Agent Iterated Language Learning
}

\newcommand*{\affaddr}[1]{#1} 
\newcommand*{\affmark}[1][*]{\textsuperscript{#1}}
\newcommand*{\email}[1]{\texttt{#1}}

\author{%
Yuchen Lian\affmark[1,2],
\ \ Arianna Bisazza\affmark[3],
\ \ Tessa Verhoef\affmark[2]\\
\affaddr{\affmark[1]Faculty of Electronic and Information Engineering, Xi'an Jiaotong University}\\
\affaddr{\affmark[2]Leiden Institute of Advanced Computer Science, Leiden University}\\
\affaddr{\affmark[3]Center for Language and Cognition, University of Groningen}\\
\email{\{y.lian, t.verhoef\}@liacs.leidenuniv.nl}\\
\email{{a.bisazza@rug.nl}}\\
}

\begin{document}
\maketitle
\begin{abstract}

Natural languages display a trade-off among different strategies to convey syntactic structure, such as word order or inflection. 
This trade-off, however, has not appeared in recent simulations of iterated language learning with neural network agents \cite{chaabouni-etal-2019-word}.
We re-evaluate this result in light of three 
factors that play an important role in comparable experiments from the Language Evolution field:
(i) speaker bias towards efficient messaging,
(ii) non systematic input languages, and
(iii) learning bottleneck.
Our simulations show that neural agents mainly strive to maintain the utterance type distribution observed during learning, instead of 
developing a more efficient or systematic language.


\end{abstract}

\section{Introduction}

The world's languages show immense variety, but linguistic patterns also show important universal tendencies \citep{greenberg1963universals}. 
It has been argued that these common design features are shaped by human cognitive constraints and pressures during communication and transmission \citep{kirby2014iterated}. 
A well-known example of these tendencies 
is the trade-off between case marking and word order as redundant strategies to encode the role of sentence constituents \cite{sinnemaki2008complexity,futrell-15-quantifying}:
flexible order typically correlates with the presence of case marking (e.g. in Russian), while fixed order is often observed in languages with little or no case marking (e.g. English).


Researchers interested in language universals and their origins have extensively used agent-based modeling techniques to study the impact of social processes on the emergence of linguistic structures \citep{de2006computer}. Besides the horizontal transmission that is often modeled in the referential game setup, the process of iterated learning, where signals are transmitted vertically from generation to generation, has been identified to shape language \cite{kirby2001spontaneous, kirby2014iterated}.

Recently, the advent of deep learning based NLP has triggered a renewed interest in agent-based simulations of language emergence. 
Most existing studies simulate the emergence of language by letting neural network agents play referential games and studying the signals they use \citep{kottur-etal-2017-natural,havrylov2017emergence,lazaridou2018emergence,chaabouni-2019-antiefficient, dagan2020co}.
By contrast,
\citet{chaabouni-etal-2019-word} 
expose their agents to a pre-defined language,
which is then learned and reproduced iteratively by a chain of agents. 
They analyze how specific properties of the initial languages affect learnability, and further investigate how agent biases affect the evolution across generations.
Among others, they studied whether neural agents tend to avoid redundant coding strategies as natural languages do.
However, the case-marking/word-order trade-off did not clearly appear in their iterated learning experiments.

\begin{figure}[ht]
  \small
    \begin{tabular}[b]{@{\ } l @{\ \ } l @{\ }}
    \hline
     \textbf{Type} &\textbf{Utterance} \\ \hline
Fix+Marker &\it m1 up 2 m2 left 3 m3 down 1 \\
\hline
Fix &\it up 2 left 3 down 1 \\
\hline
 &\it m1 up 2 m3 down 1 m2 left 3 \\
Free-order &\it m2 left 3 m1 up 2 m3 down 1 \\
+Marker &\it m3 down 1 m1 up 2 m2 left 3 \\
 &  \ \ \ \ \ \ \ \ \ \ \ \ \ \ \ \ \ \ \  ... \\
\hline
\end{tabular}
\includegraphics[width=.275\columnwidth]{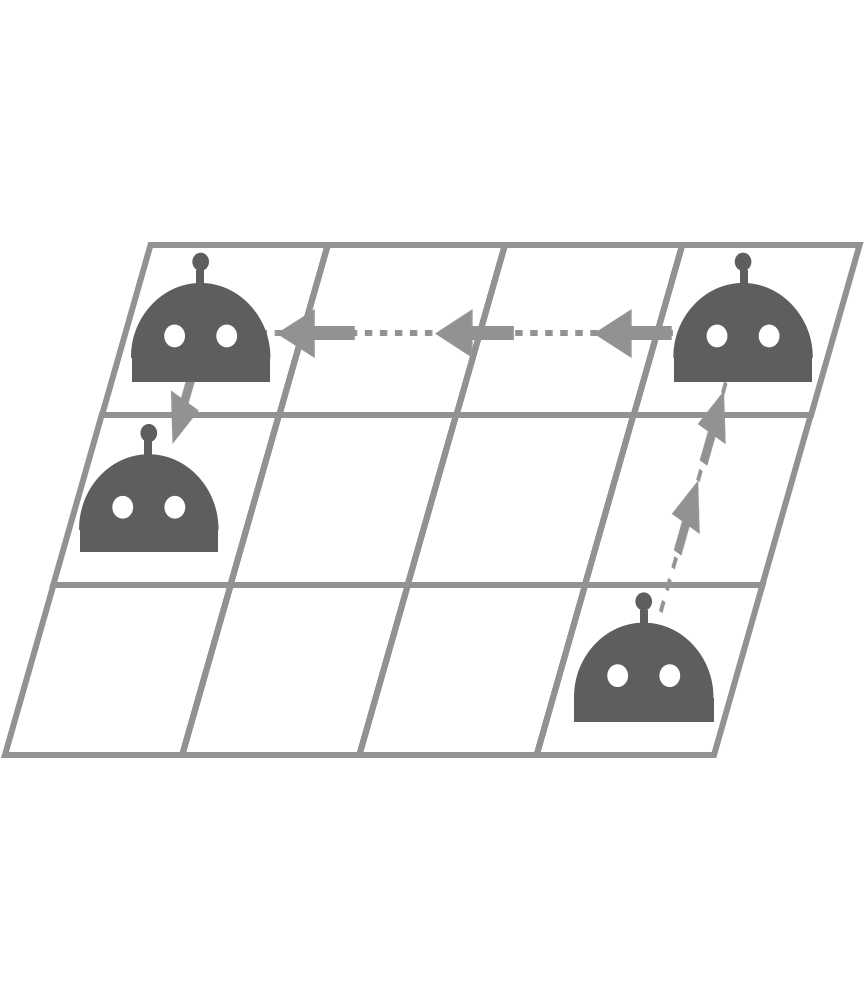}
\caption{\label{Tbl:Example_language}
Utterances corresponding to \textsc{`up up~left left left down'}, in three basic languages.}
\end{figure}

In this work, we re-evaluate this finding in light of three factors that play an important role in comparable experiments from the Language Evolution field:
(i) speaker bias towards efficient messaging \citep{i2003least},
(ii) unpredictable variation in the initial languages \citep{smith2010eliminating,fedzechkina2017balancing},  (iii) exposure to a limited set of example utterances, known as
`learning bottleneck' \citep{kirby2014iterated}.
We follow the iterated learning setup of \newcite{chaabouni-etal-2019-word} 
where neural agents are trained to communicate about trajectories in a simple gridworld, exchanging instructions in miniature languages (Fig.~\ref{Tbl:Example_language}).%
\footnote{The original implementation is taken from \url{https://github.com/ facebookresearch/brica}.
Our revised code and data are available at \url{https://github.com/Yuchen-Lian/neural\_agent\_trade-off}.}

\section{Miniature Languages}
\label{mini_lang}

Word order and case marking are two different mechanisms to convey sentence constituent roles, both widely attested among world languages. 
Both have been shown to serve as equally valuable cues to learn grammatical roles during language learning by children \cite{slobin1982}, and by simple recurrent neural networks \cite{lupyan2002case}. 
%
To model these mechanisms we use simple artificial languages based on \newcite{chaabouni-etal-2019-word}.
The meaning space is composed of trajectories defined by random combinations of four oriented actions $\{$\textsc{left,\ right,\ up,\ down}$\}$.
Each utterance or sentence (S) consists of several phrases (P), which in turn are composed of a command (C) and a quantifier (Q):
\begin{align}
    \label{rule1} &S \to P_i P_j P_k ...\\
    \label{rule2} &P_i | P_j | P_k | ... \to C Q \\
    \label{rule3} &C \to (left|right|up|down) \\
    \label{rule4} &Q \to (1|2|3)
\end{align}
where $left, right, up, down, 1, 2, 3$ are spoken words which are atomic elements of the language.
%
We consider three basic language types:  Fixed-order with marker (redundant), Fixed-order without marker (non-redundant), and Free-order with marker (non-redundant). See examples in
Fig.~\ref{Tbl:Example_language}. 

\paragraph{Fixed-order vs. Free-order}
This concerns Rule~\ref{rule1}: In a fixed-order language, the order of phrases strictly corresponds to the temporal order of instructions in the trajectory.\footnote{This is the `forward-iconic' language of \newcite{chaabouni-etal-2019-word}. We do not consider other fixed orders in this work, as we are mostly interested in the contrast between redundant and non-redundant languages.}
Free-order languages, instead, allow any permutation of phrases.
For instance, there are in total six possible free-order utterances for a 3-phrase trajectory (Fig.~\ref{Tbl:Example_language}).

\paragraph{Case Marking}
In a case-marking language, each phrase is preceded by a temporal marker indicating its role.
Thus, Rule~\ref{rule2} changes to: $P_i \to m_i\ C\ Q$ with the marker $m_i$ indicating that $CQ$ is the $i^{th}$ action segment.
Note that a fully free-order language is unintelligible without markers. 



\section{Neural-Agents Iterated Learning}
\label{agents}

We strictly follow the iterated learning setup of \newcite{chaabouni-etal-2019-word} unless explicitly noted. 

\paragraph{Agent architecture}
Agents are implemented as 1-layer attentional Seq2Seq \citep{NIPS2014_a14ac55a,bahdanau2014neural} LSTM \citep{hochreiter1997long} networks.
Each agent acts as both speaker (receiving trajectories and describing them with utterances) and listener (receiving utterances and trying to induce the corresponding trajectories).
The vocabulary contains both actions and words; embeddings of the encoder input and decoder output are tied \cite{press-wolf-2017}. 

\paragraph{Individual and iterated learning}
Given trajectory-utterance pairs, agents are trained by teacher forcing \citep{goodfellow2016deep} in both listening and speaking mode, using the early-stopping and optimizer settings of \citet{chaabouni-etal-2019-word}.\footnote{To handle 1-to-N trajectory-to-utterance mappings in free-order languages,
\newcite{chaabouni-etal-2019-word} used a modified training loss for the Speaker direction. 
Empirically, we find that sampling multiple free-order utterances in the initial training corpus leads to very similar results, so we do not use the modified training loss. This allows us to support more complex languages without major changes to the training procedure.}
Iterated learning \citep{kirby2001spontaneous} is implemented by letting a trained adult agent teach a randomly initialized child agent, and repeating this process for a number of generations. 
At each generation, two steps are performed: 
(1) a trained adult agent receives a batch of trajectories and generates utterances by sampling from its own decoder outputs;
(2) a randomly initialized child agent is trained on these agent-specific trajectory-utterance pairs. 
As an exception, the generation-0 agent is directly trained on the corpus generated by a given miniature grammar.

\paragraph{Evaluation}
\label{sec:evaluation}

In both speaking and listening mode, sequences are generated by greedy decoding and evaluated by sentence-level accuracy.
Listener evaluation is standard as the true meaning of an utterance 
is unique. 
For speakers in the first generation, instead, we consider all acceptable utterances according to the grammar as candidate targets;
for later generations, we take $k = i!$ utterances sampled from the parent's speaking network as targets ($i$ is the maximum number of phrases per trajectory).
Validation for early stopping is performed similarly.
These evaluation procedures allow a child agent's language to deviate from the parent language, even while achieving perfect accuracy. 



For each experiment, we report speaking accuracy, listening accuracy, as well as average utterance length across generations. 
To get more insight into language changes, we plot the distribution of utterance types in the adult speaking agents across generations.
Specifically, we count how often an utterance belongs to one of the basic language types (\textsl{fix, fix\_marker, free, free\_marker}), and how often markers are dropped for some of the phrases (\textsl{fix\_drop, free\_drop}). Utterances that do not fall into any of these categories are labeled as \textsl{`other'}.

For more technical details on the agents and evaluation, see Appendix \ref{app:Training_details}.
Example utterances at various generations are shown in Appendix \ref{app:uttr_table}.



\begin{figure*}[ht]
     \centering
     \begin{subfigure}[b]{0.245\textwidth}
         \centering
         \includegraphics[width=\textwidth]{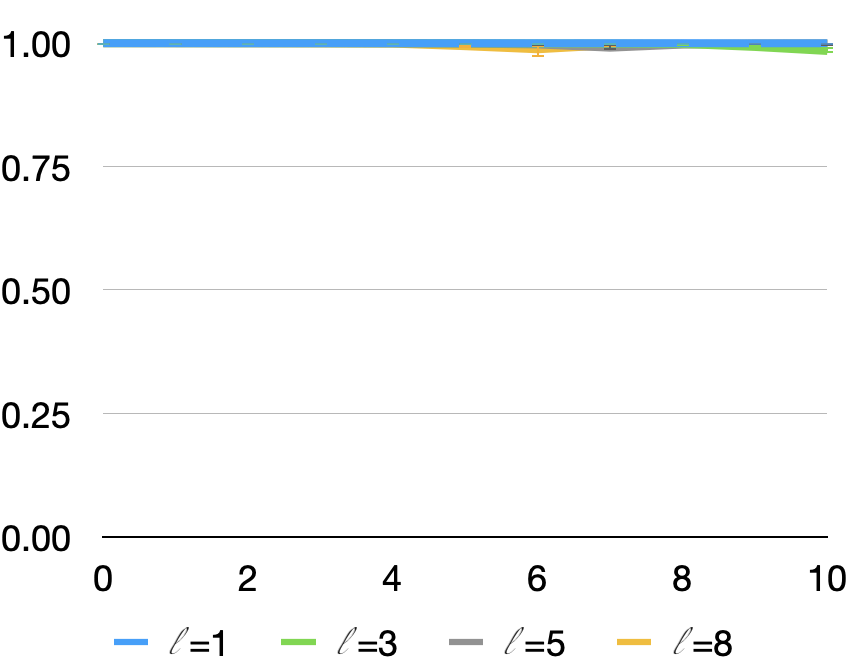}
         \caption{Speaking accuracy}
         \label{fig:fix-spk}
     \end{subfigure}%
     \hfill
     \begin{subfigure}[b]{0.245\textwidth}
         \centering
         \includegraphics[width=\textwidth]{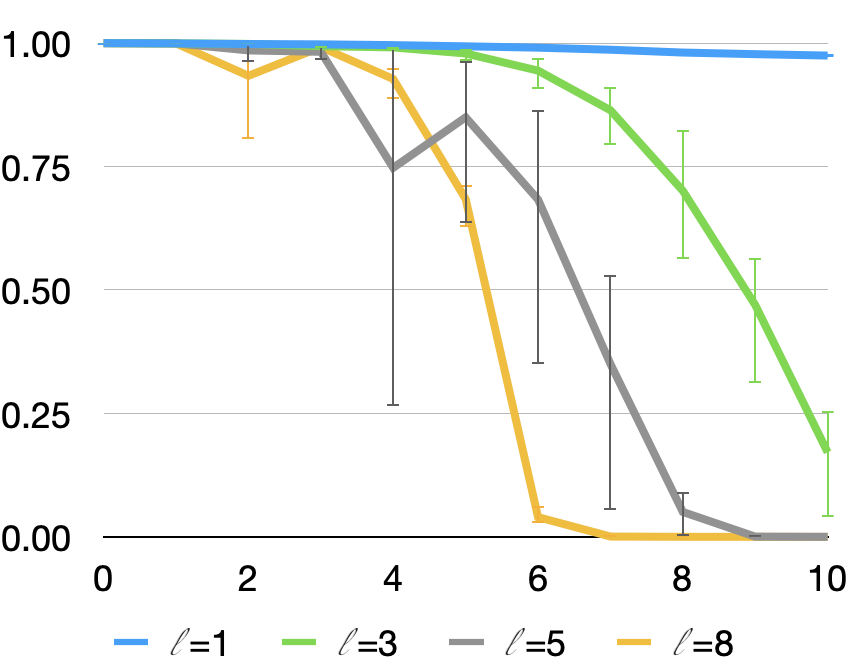}
         \caption{Listening accuracy}
         \label{fig:fix-lis}
     \end{subfigure}%
     \hfill
     \begin{subfigure}[b]{0.25\textwidth}
         \centering
         \includegraphics[width=\textwidth]{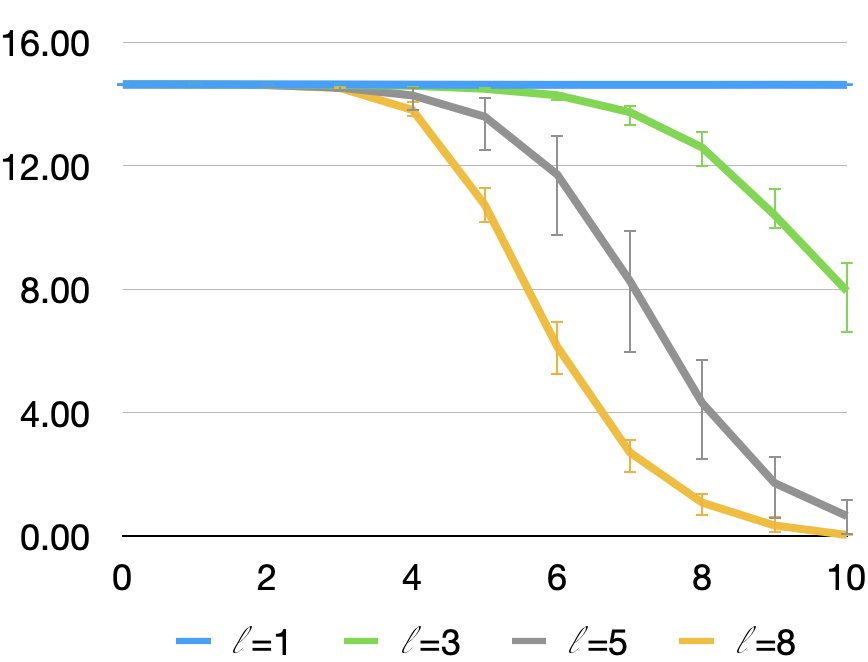}
         \caption{Average utterance length}
         \label{fig:fix-len}
     \end{subfigure}%
     \hfill
     \begin{subfigure}[b]{0.26\textwidth}
         \centering
         \includegraphics[width=\textwidth]{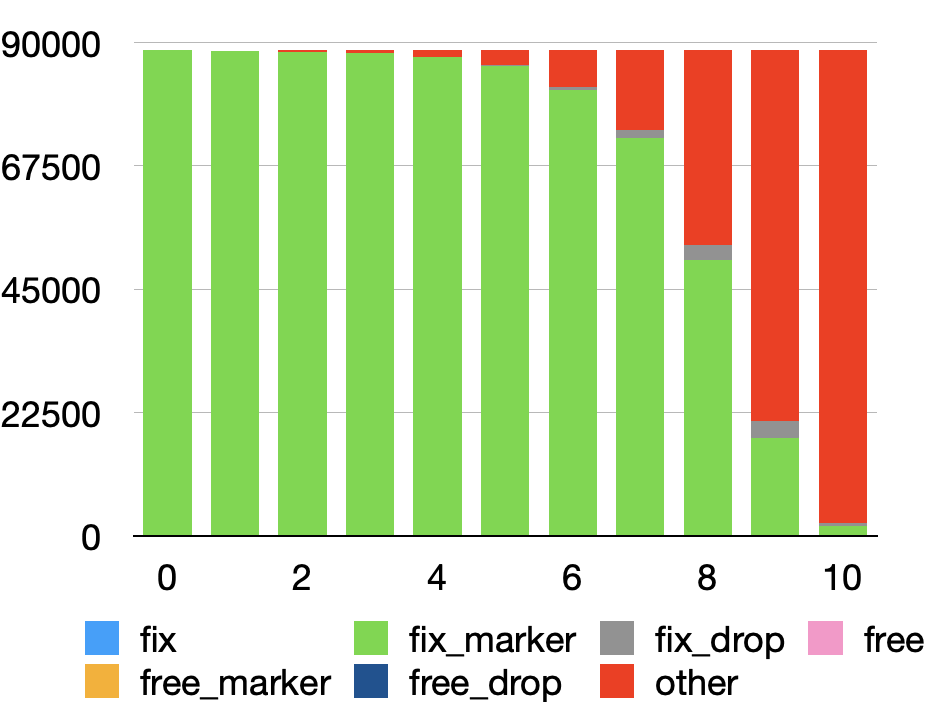}
         \caption{Utterance types ($\ell=3$)}
         \label{fig:fix-ana}
     \end{subfigure}%
        \caption{Iterated learning of the Fixed+Marker language with least-effort pressure of varying strength ($\ell$), over 10 generations. Results in (a,b,c) are averaged over three random-seed initializations. 
        (d) shows the distribution of utterance types of the speaking adult agents with $\ell = 3$ (one seed only). 
        }
        \label{fig:forward_shortselection}
\end{figure*}

\section{Effect of Least-Effort Bias}
\label{exp1}

A bias towards efficient messaging has been proposed as explaining factor for several tendencies observed in natural languages \citep{i2003least, kanwal2017zipf, fedzechkina2017balancing}.
Could the lack of such a bias in neural networks explain the survival of redundant languages? 
To verify this, we design a simple mechanism simulating an agent's preference to minimize utterance length, based on \citet{chaabouni-etal-2019-word}'s framework:
To teach the next generation, given a trajectory $\mathbf{t}$, an adult agent generates $n$ (possibly identical) utterances 
$\{ \hat{u} \} = \{ \hat{u_1}, \hat{u_2}, ... \hat{u_n} \} $
by sampling from its trained network.
Instead of modifying the training process, we exploit the diversity occurring in the sampled utterances and hard-code 
a shorter-sentence selection bias into this adult language generation. 
As shown in Algorithm~\ref{algo}, the sampling function is called $n$ times to generate $n$ samples.
In turn, at each iteration, we ask the adult speaker to generate $\ell$ sentences and select each time the shortest one.
Thus, we can control the bias strength by varying the number of generated samples ($\ell$) in each of the $n$ iterations.
As $\ell$ increases, the chances of sampling a shorter sentence increase, resulting in a stronger pressure; $\ell=1$ means no bias.
We expect this least-effort bias will cause the redundant disambiguation mechanism to gradually disappear, 
and the fixed-order strategy to dominate as that always leads to shorter utterances.


\setlength{\textfloatsep}{4pt}
\begin{algorithm}[h]
\small
\caption{Shorter-sentence selection}
\label{algo}
\LinesNumbered
\KwIn{Trajectory $t$}
\KwOut{$n$ sampled utterances $\{\hat{u}\}$}
\For{$j=1:n$}{
    \eIf{$shorter\_selection$}{
        $uttrs$ = Adult.speaker($t$).sample($\ell$)\\
        $uttr\_select$ = $uttrs[0]$ \\
        \For{$i=1:\ell$}{
            $u$ = $uttrs[i]$ \\
            \If{$len(u) \le len($uttr\_select$)$}{
                $uttr\_select$ = $u$ \\
            }
        }
    }{$uttr\_select$ = Adult.speaker($t$).sample(1)}
    {$\{\hat{u}\}$.append($uttr\_select$)}
}
\end{algorithm}

\paragraph{Results}
Fig.~\ref{fig:forward_shortselection} shows the iterated learning results of the Fixed+Marker language with various levels of least-effort bias $\ell= \{ 1, 3, 5, 8\}$, which represent no pressure, low- , medium- and high-level pressure towards shorter utterances, respectively.
The experiment without least-effort pressure ($\ell=1$) corresponds to the setup of \newcite{chaabouni-etal-2019-word}, in which the redundant language was found to remain stable across generations.

We find that, while speaking accuracy remains stable (\ref{fig:fix-spk}), our least-effort pressure leads to a severe drop in listening accuracy  (\ref{fig:fix-lis}) and a dramatic increase of \textsl{other} types in the speaking adult agent starting from the fifth generation (\ref{fig:fix-ana}).  
Stronger levels of pressure lead to a faster decrease of average utterance length (\ref{fig:fix-len}), which was expected.
However, manual inspection of the utterances (Appendix \ref{app:uttr_table}) reveals that agents start dropping entire phrases, thereby losing information, instead of either dropping markers or changing the word order.

\section{Effect of Input Language Variability}
\label{exp2}

In the languages of \citet{chaabouni-etal-2019-word}, 
markers are either present and fully systematic, or not present at all. If there is no marker example in the initial language, it is unlikely an agent would suddenly invent it. Conversely, a fully systematic use of markers may be perfectly learnable by the agent, unlikely to change or disappear over generations. 
Unpredictable variation, instead, is a common feature of the artificial languages used in human learning studies.
For example \newcite{fedzechkina2017balancing} combine \textit{optional} case marking with either fixed or free word order, 
while \newcite{smith2010eliminating} use two plural markers with different distributions over nouns.
Inspired by this, we experiment with unpredictable variations in the use of markers, namely: (i) variability \textit{among} utterances, where each utterance is consistent with one of the basic language types chosen at random, 
and (ii) variability \textit{within} utterances, where the use of markers is also unpredictable within a single utterance.

\COMMENT{
\renewcommand{\arraystretch}{1.1}
\begin{table}[ht]
\centering \small
\begin{tabular}
{@{\ } l @{\ \ \ \ \ } l @{\ }}
\hline
Mix & Mix\_drop \\ \hline
\it m1 up 2 m2 left 3 m3 down 1 & \it m1 up 2 left 3 down 1 \\
\it m1 up 2 m2 left 3 m3 down 1 & \it m1 up 2 m2 left 3 m3 down 1 \\
\it up 2 left 3 down 1  &  \it up 2 m2 left 3 m3 down 1\\
\it up 2 left 3 down 1  &  \it m2 left 3 m1 up 2 down 1\\
\it m2 left 3 m1 up 2 m3 down 1 &  \it \it down 1 m2 left 3 m1 up 2\\
\it m3 down 1 m2 left 3 m1 up 2 &  \it left 3 down 1 up 2\\
\hline
\end{tabular}
\caption{\label{Tbl:Example_mix} Example utterances for trajectory \textsc{\small{`up up~left left left down'}} in Mix and Mix\_drop language.}
\end{table}
}

\subsection{Variability Among Utterances}
\label{mix-learning}

For every trajectory in the initial training set, two utterances are generated for each of the three basic language types:
(i) Fixed-order+Marker,
(ii) Fixed-order without markers and
(iii) Free-order+Marker. 
%
Our goal here is to find out whether the agents will tend to prefer any of the three language types over generations, according to their inherent biases.

\begin{figure*}[tb]
     \centering
     \begin{subfigure}[b]{0.25\textwidth}
         \centering
         \includegraphics[width=\textwidth]{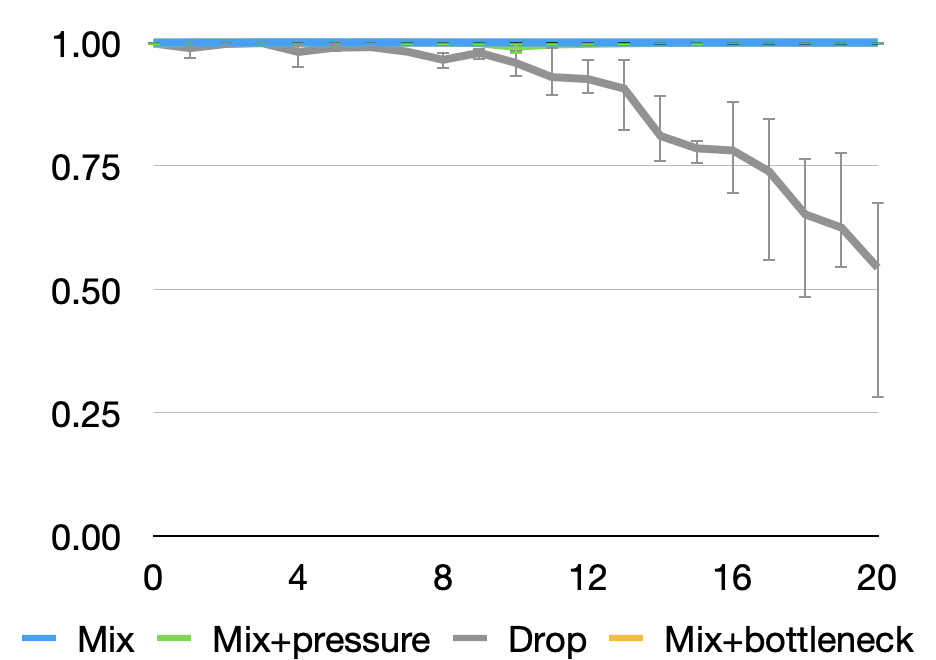}
         \caption{Speaking accuracy}
         \label{fig:2_spk}
     \end{subfigure}%
     \hspace{1em}%
     \begin{subfigure}[b]{0.25\textwidth}
         \centering
         \includegraphics[width=\textwidth]{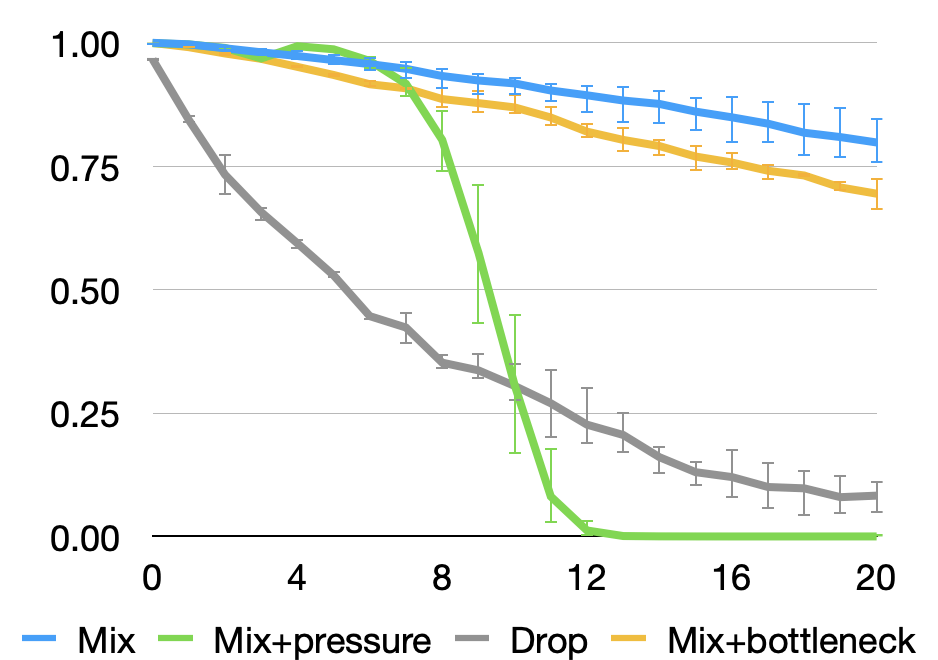}
         \caption{Listening accuracy}
         \label{fig:2_lis}
     \end{subfigure}%
     \hspace{1em}%
     \begin{subfigure}[b]{0.25\textwidth}
        \centering
        \includegraphics[width=\textwidth]{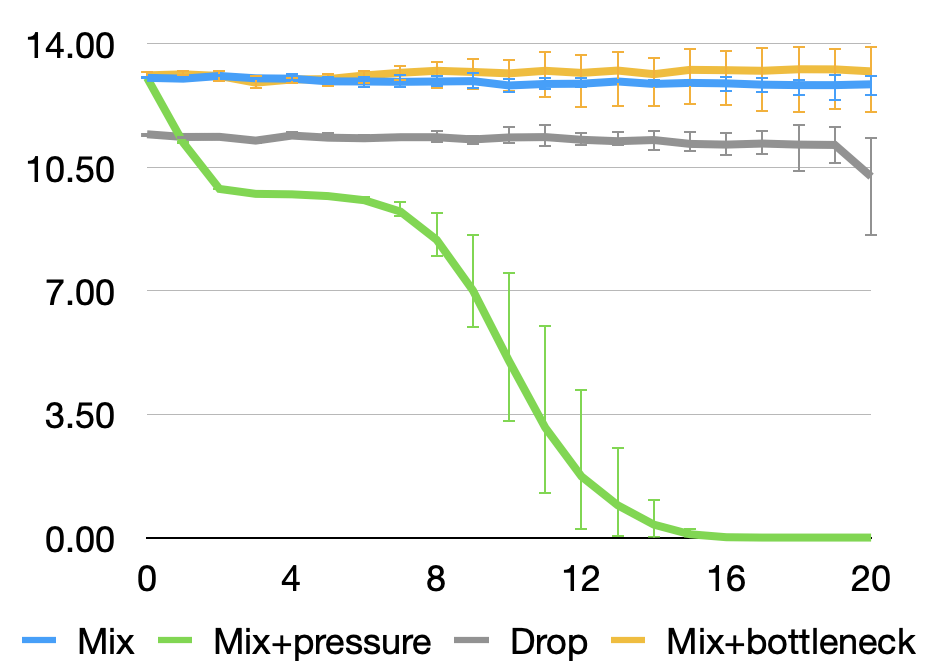}
         \caption{Average utterance length}
         \label{fig:2_len}
     \end{subfigure}%
     \\
     \begin{subfigure}[b]{0.25\textwidth}
         \centering
         \includegraphics[width=\textwidth]{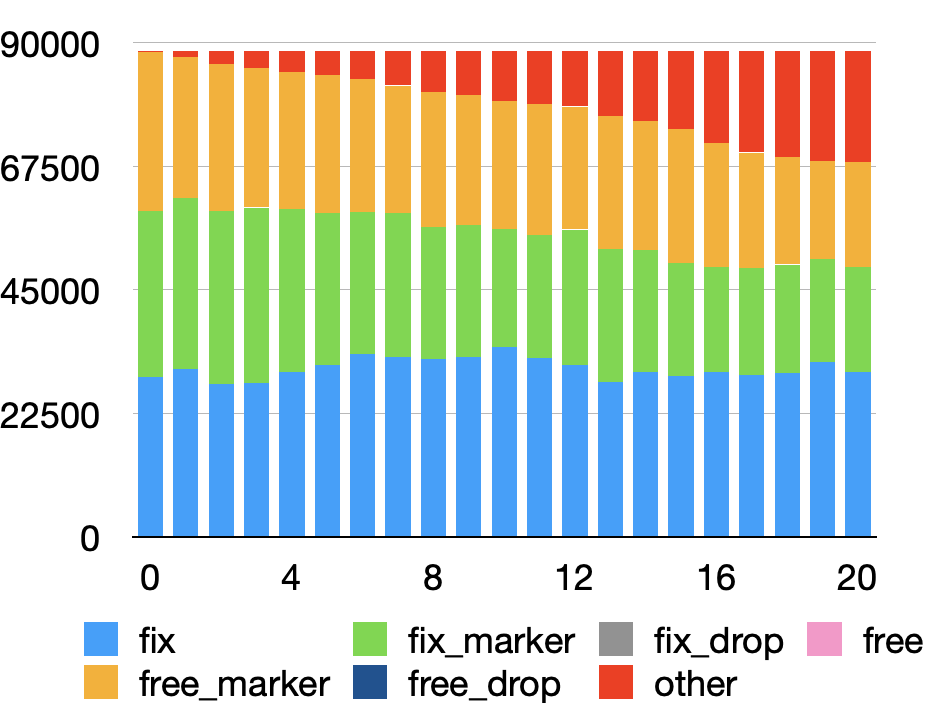}
         \caption{Utterance types (Mix)}
         \label{fig:ana_mix}
     \end{subfigure}%
     \hfill
     \begin{subfigure}[b]{0.25\textwidth}
        \centering
        \includegraphics[width=\textwidth]{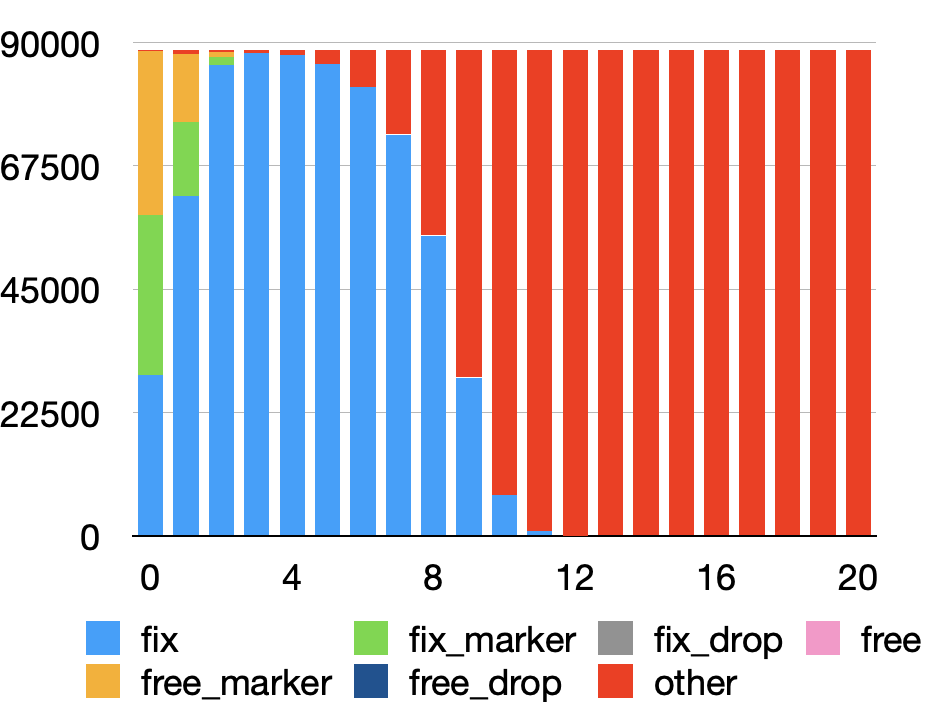}
         \caption{Utter. types (Mix+pressure)}
         \label{fig:ana_mix-shot}
     \end{subfigure}%
     \hfill
     \begin{subfigure}[b]{0.25\textwidth}
         \centering
         \includegraphics[width=\textwidth]{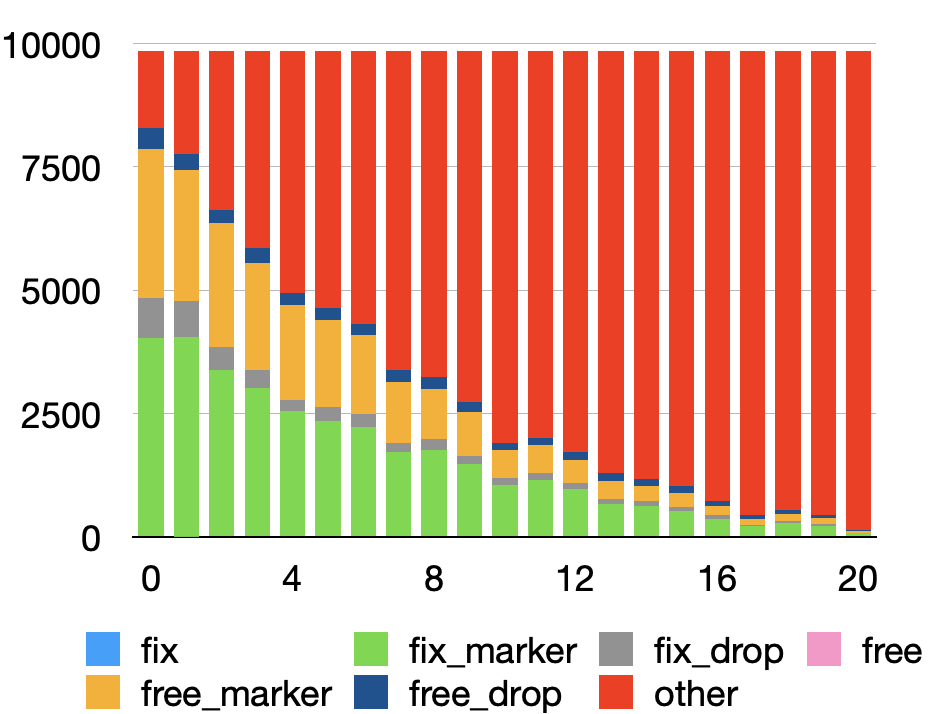}
         \caption{Utter. types (Mix\_drop)}
         \label{fig:ana_drop}
     \end{subfigure}%
     \hfill
     \begin{subfigure}[b]{0.25\textwidth}
         \centering
         \includegraphics[width=\textwidth]{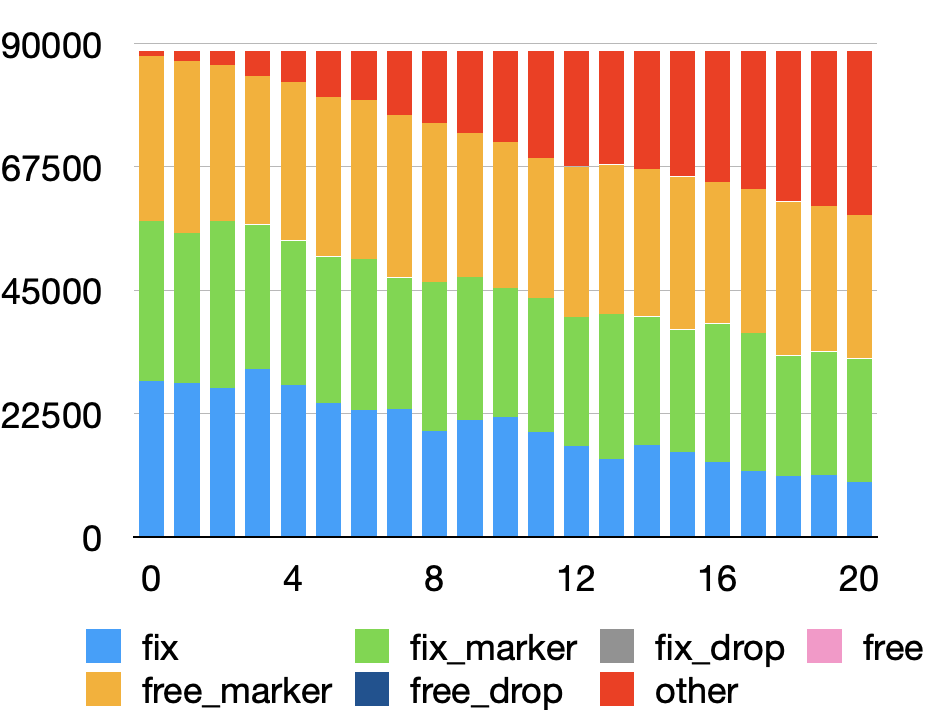}
         \caption{Utter. types (Bottleneck)}
         \label{fig:ana_split}
     \end{subfigure}%
        \caption{
        Mixed language learning
        without and with least-effort pressure, drop-marker language without least-effort pressure, and mixed language with learning bottleneck. Results in (a,b,c) are averaged over three random seeds.
        (d,e,f,g) show the respective distributions of utterance types in the speaking adult agents (one seed only).%
        \footnote{* The speaking accuracies for Mix+pressure, Drop and Mix+bottleneck in (a) are not visible because they are similar to the accuracy of Mix (blue line).}
        }
    \label{fig:2}
\end{figure*}

\paragraph{Results}
\label{mix-result}
%
Results are shown in Fig.~\ref{fig:2} (blue lines).
The overall high speaking accuracy suggests that
the agents can learn to imitate their parents' language very well. 
We observe a slow, but steady, loss in listening accuracy, which we attribute to the random sampling errors from the parent speaker and the natural presence of errors in the neural network learning process.
%
Besides a steady increase of uncategorizable utterances (\textsl{other}) in Fig.~\ref{fig:ana_mix}, the distribution of the three language types remains relatively stable even after 20 generations.
We looked for sentences where only some of the markers are dropped (\textsl{free\_drop/fix\_drop}) but found almost none. See examples in Appendix \ref{app:uttr_table}.
These results show that mixing language types in the initial training set is not sufficient to induce the loss of redundant encoding with agent learners, as opposed to human learners \cite{fedzechkina2017balancing}.


\paragraph{Results with Least-Effort Bias}
We also study the combination of the two factors (Mix+pressure) using medium pressure ($\ell=3$). This setup (Fig.~\ref{fig:2}, green lines)
leads to a more efficient language during the first five generations, as shown by the initially stable speaking and listening accuracy and a decrease of average length.
This phase corresponds to a proliferation of fixed-order no-marker sentences and the disappearance of the other language types (\ref{fig:ana_mix-shot}).
In only two generations, 
this language has reached the shortest possible overall length while remaining intelligible.
After a while, however, 
child agents start to 
receive shorter but incorrect utterances, resulting in a drop of listening accuracy and, finally, an unintelligible language. 


\subsection{Variability Within Utterances}
\label{sect:dropmarker}

We design a more unpredictable language where each phrase marker is randomly dropped according to a given probability (10\%). 
Half of the utterances are fixed- and half are free-order. See examples in Appendix \ref{app:uttr_table}.
This language is closely inspired by those of \newcite{fedzechkina2017balancing}.
We expect agents will either stop using markers completely over generations, or use them more consistently.



\paragraph{Results}

Despite the relatively small probability of dropping a marker, speaking and listening accuracies drop rapidly (\ref{fig:2_spk} and \ref{fig:2_lis}, grey lines) and
\textsl{other} utterances become dominant (Fig.~\ref{fig:ana_drop}).
This language becomes unintelligible before regularization is observed, again challenging our expectations.


\section{Effect of Learning Bottleneck}\label{bottleneck}

Although real languages support the production of enormous sets of utterances, human learners master them by exposure to a limited number of examples. This \textit{learning bottleneck} acts as a pressure forcing languages to regularize during cultural transmission \citep{smith2003iterated, brighton2005language}.
Human-based experiments and computational simulations have found that 
this pressure can lead to increased structure in emerging language systems \citep{kirby2014iterated}. 
We apply a learning bottleneck to our mixed language experiment (Sect.~\ref{mix-learning}) by randomly sub-sampling, at each iteration,  50\% of the data used to train the next generation. 
Evaluation and training details are the same as in Sect.~\ref{mix-learning}.

\paragraph{Results}
We find that training data sub-sampling leads to a slightly steeper drop in listening accuracy (yellow \textit{vs.} blue line in Fig.~\ref{fig:2_lis}), 
but the respective distributions of utterance types (Fig.~\ref{fig:ana_split} vs. \ref{fig:ana_mix})
remain very similar, suggesting the learning bottleneck does not result in a more structured language.

\footnotetext{Speaking accuracies of Mix+pressure, Mix+bottleneck, Drop
in \ref{fig:2_spk} are hidden behind the accuracy of Mix (blue line).}

\section{Discussion and Conclusions}
Neural-agent iterating learning is a promising framework to study the impact of social processes on the emergence of patterns and language universals, like the word order/case marking trade-off. 
However, previous work with LSTM-based agents \cite{chaabouni-etal-2019-word} has failed to replicate this human-like pattern.
We re-evaluated this finding by (i)~hard-coding a least-effort bias into our agents, (ii)~designing less systematic input languages, 
and
(iii)~introducing a learning bottleneck.
In all cases, our agents proved to be accurate learners, but the patterns of language change over generations did not match our expectations.
Specifically, least-effort bias (\S\ref{exp1}) and highly unpredictable input language (\S\ref{sect:dropmarker}) led to communication failure, whereas moderate input language variability (\S\ref{mix-learning}) and learning bottleneck (\S\ref{bottleneck}) led to a stable language distribution, confirming previous observations on the survival of redundant coding strategies in neural-agent iterated learning \cite{chaabouni-etal-2019-word}. 
Only combining least-effort bias with moderate language variability (\S\ref{mix-learning}) led to a temporary optimization of the language, but that was again followed by communication failure.




In real language use, the pressure to reduce effort is balanced with communicative needs \cite{kirby2015compression, regier201511}
and does normally not lead to severe language degradation. 
Future work should design subtler least-effort biases, for instance considering efficiency in terms of grammatical structure and cognitive effort. 
Moreover, our results with non fully systematic
languages show that 
agents strive to preserve the initial  utterance type distribution. 
In human learning, this behavior is called probability matching 
and 
is affected by task complexity: more difficult tasks tend to regularization or over-matching \citep{ferdinand2019cognitive,kam2009getting}, where the more frequent variant is chosen more often than it appeared in the input.
Over many generations, even a slight over-matching can lead 
to the emergence of linguistic regularities \citep{smith2010eliminating}
like the word order/case marking trade-off observed in human learners by \citet{fedzechkina2017balancing}. 
%
%
We conclude that  the  current neural-agent iterated learning framework  is  not  yet  ready to simulate language evolution processes in a human-like way.
More natural cognitive biases supporting efficiency should be modeled, while the speaker objective should be balanced with a measure of communicative success, such as the likelihood of a message to be understood by the listener 
\cite{goodman2016pragmatic,tessler21}.

\section*{Acknowledgements}

Lian was partly funded by the China Scholarship Council (CSC 201906280463).
Bisazza was partly funded by the Netherlands Organization for Scientific Research (NWO project 639021646).

\bibliography{anthology,new_bib}
\bibliographystyle{acl_natbib}

\clearpage
\appendix

\section{Training details}
\label{app:Training_details}
Following \newcite{chaabouni-etal-2019-word} we limit the maximum number of segments per trajectory, $i$, to 5 and at most 3 steps per phrase, resulting in a total of 89k possible trajectories. 
Subsequently, the number of candidate target utterances during evaluation is set to $k=i!=120$.
As an exception, for the drop-marker language (Section~\ref{sect:dropmarker}) we limit $i$ to 4 instead of 5 due to the computational cost of enumerating all correct utterances for a trajectory in this language during validation (accordingly, $k$ is reduced to 24).
The trajectory-utterance pairs are randomly split into training, validation and test sets with a proportion of 80\%, 10\% and 10\% respectively. 

We fix the hidden layer size (20) and batch size (16) for all experiments.
Similar to \newcite{chaabouni-etal-2019-word}, we use the Amsgrad optimizer \cite{j.2018on}. For each generation, the maximum number of training epochs is set to 100 and we stop the training if both speaking and listening accuracy on development set have no improvement over 5 epochs.
To ensure the reliability of our results, we repeat each experiment with 3 different random seeds and observe trends over 20 generations (unless trends are already very clear after 10, as in Fig.~\ref{fig:forward_shortselection}). 




\section{Example utterances at various generations}
\label{app:uttr_table}
We let each agent generate six utterances corresponding to the trajectory \textsc{`right up up down right right right'}. As some of these utterances are identical, we remove the duplicates and only list unique ones in Tab. \ref{Tbl:agent_uttr_samples}.

\renewcommand{\figurename}{Table}
\addtocounter{figure}{-1}
\begin{figure*}[b]

\resizebox{\textwidth}{!}{%
\begin{tabular}
{l l l l}
\hline
 & Fix+Marker with pressure ($\ell=3$) & Mix & Mix with pressure ($\ell=3$)\\ \hline
Input & M1 right 1 M2 up 2 M3 down 1 M4 right 3
        & right 1 up 2 down 1 right 3 
        & right 1 up 2 down 1 right 3\\
        & 
        & M1 right 1 M2 up 2 M3 down 1 M4 right 3 
        & M1 right 1 M2 up 2 M3 down 1 M4 right 3 \\
        &
        & M1 right 1 M2 up 2 M4 right 3 M3 down 1
        & M2 up 2 M4 right 3 M3 down 1 M1 right 1 \\
        &
        & M3 down 1 M1 right 1 M4 right 3 M2 up 2
        & M4 right 3 M2 up 2 M3 down 1 M1 right 1 \\
        \hline
Iter\_0 & M1 right 1 M2 up 2 M3 down 1 M4 right 3
        & right 1 up 2 down 1 right 3 
        & right 1 up 2 down 1 right 3\\
        & 
        & M1 right 1 M2 up 2 M3 down 1 M4 right 3 
        & M1 right 1 M2 up 2 M3 down 1 M4 right 3 \\
        &
        & M1 right 1 M2 up 2 M4 right 3 M3 down 1  
        & M1 right 1 M2 up 2 M4 right 3 M3 down 1 \\
        \hline
Iter\_1 & M1 right 1 M2 up 2 M3 down 1 M4 right 3
        & right 1 up 2 down 1 right 3
        & right 1 up 2 down 1 right 3\\
        &
        & M1 right 1 M2 up 2 M3 down 1 M4 right 3
        & M1 right 1 M2 up 2 M3 down 1 M4 right 3\\
        & 
        & M2 up 2 M1 right 1 M4 right 3 M3 down 1
        & M3 down 1 M2 up 2 M4 right 3 M1 right 1\\
        &
        & M3 down 1 M2 up 2 M4 right 3 M1 right 1
        & \\
        \hline
Iter\_5 & M1 right 1 M2 up 2 M3 down 1 M4 right 3 M5 right 3
        & M3 down 1 M4 right 3 M2 up 2 M1 right 1
        & right 1 up 2 down 1 right 3\\
        & M1 right 1 M2 up 2 M3 down 1 M4 right 3
        & M2 up 2 M3 down 1 M4 right 3 M1 right 1 
        & \\
        & M1 right 1 M2 up M3 down 1 M4 right 3 M5 3
        & right 1 up 2 down 1 right 3
        & \\
        &
        & M1 right 1 M4 right 3 M3 down 1 M2 up 2
        & \\
        & 
        & M1 right 1 M2 up 2 M3 down 1 M4 right 3
        & \\
        & 
        & M4 right 3 M3 down 1 M1 right 1 M2 up 2
        & \\
        \hline
Iter\_10 & M1 right 1 M2 up 2 M3 down 1
        & right 1 up 2 down 1 right 3
        & right 1 up 2 down 1 right 3\\
        & M1 right 1 M2 up 2
        & M2 up 2 M3 down 1 M4 right 3 M1 right 1
        & right 1 up 2\\
        & M1 right 1
        & M1 right 1 M3 down 1 M4 right 3 M2 up 2
        & right 1\\
        & 
        & M1 right 1 M3 down 1 M4 right 3 M2 up 2
        & \\
        & 
        & right 1 up 2 down 1 right 3
        & \\
        \hline
\end{tabular}%
}

\vspace{10mm}

\resizebox{.73\textwidth}{!}{%
\begin{tabular}
{l l l}
\hline
 & Mix\_drop & Mix with learning bottleneck\\ \hline
Input   & M1 right 1 M2 up 2 M3 down 1 M4 right 3
        & right 1 up 2 down 1 right 3\\
        & M1 right 1 M2 up 2 down 1 M4 right 3
        & M1 right 1 M2 up 2 M3 down 1 M4 right 3 \\
        & M2 up 2 right 1 M4 right 3 M3 down 1
        & M3 down 1 M1 right 1 M2 up 2 M4 right 3\\
        & down 1 M1 right 1 up 2 M4 right 3
        & M4 right 3 M2 up 2 M3 down 1 M1 right 1 \\
        & M1 right 1 M4 right 3 M2 up 2 M3 down 1
        & \\
        \hline
Iter\_0 & right 3 M2 up 2 M3 down 1 M4 right 3
        & right 1 up 2 down 1 right 3 \\
        & M1 right 1 up 1 M3 down 1 M2 up 2 M4 right 3  
        & M3 down 1 M1 right 1 M2 up 2 M4 right 3 \\
        & M1 right 1 M4 right 3 M3 down 1 M2 up 2  
        & M1 right 1 M2 up 2 M3 down 1 M4 right 3 \\
        & M1 right 1 M2 up 2 M3 down 1 M4 right 3
        & \\
        & M3 down 1 M1 right 1 M4 right 3 M2 up 2
        & \\
        \hline
Iter\_1 & M3 down 1 M1 right 1 M2 up 2 M4 right 3
        & M3 down 1 M4 right 3 M2 up 2 M1 right 1\\
        & M2 up 2 M4 right 3 M1 right 1 M3 down 1
        & M4 right 3 M3 down 1 M2 up 2 M1 right 1\\
        & M2 up 2 M1 right 1 M3 down 1 M4 right 3
        & M4 right 3 M3 down 1 M1 right 1 M2 up 2\\
        & M1 right 1 M2 up 2 M3 down 1 M4 right 3
        & right 1 up 2 down 1 right 3\\
        &
        & M1 right 1 M2 up 2 M3 down 1 M4 right 3\\
        \hline
Iter\_5 & M3 down 1 right 3 M1 right 1 M2 up 2
        & M2 up 2 M4 right 3 M1 right 1 M5 right 3\\
        & M2 up 2 M1 right 1 M3 down 1 M4 right 3
        & M2 up 2 M3 down 1 M4 right 3 M1 right 1\\
        & right 2 M2 up 2 M3 down 1 M4 right 3
        & right 1 up 2 down 1 right 3\\
        & M1 right 1 M2 up 3 M3 down 1 M4 right 3 
        & M1 right 1 M3 down 1 M4 right 3 M2 up 2\\
        & M1 right 1 M2 up 2 M3 down 1 M4 right 3
        & \\
        & M3 down 1 right 3 M4 right 3 M1 right 1
        & \\
        \hline
Iter\_10 & M1 right 1 M3 down 1 M4 right 3 M1 right 1
        & M1 right 1 M2 up 2 M3 down 1 M4 right 3\\
        & M4 right 3 down 1 M1 right 1 M2 down 1 down 1
        & \\
        & M4 right 3 M2 up 2 M4 right 3
        & \\
        & M1 right 2 M2 up 3 M3 down 1 M4 right 3
        & \\
        & M3 down 1 M1 right 2 M4 right 3 M2 up 1
        & \\
        & M1 right 1 M3 down 2 M4 right 3 M3 down 1
        & \\
        \hline
\end{tabular}%
}

\caption{\label{Tbl:agent_uttr_samples} Utterances sampled from the agents' speaking network given the trajectory \textbf{\textsc{`right up up down right right right'}} in Fix+Marker language learning with pressure (\S\ref{exp1}), Mix language learning without and with pressure (\S\ref{mix-learning}), Mix\_drop language learning (\S\ref{sect:dropmarker}) and Mix language with learning bottleneck (\S\ref{bottleneck}).
For each experiment and each generation, we show six randomly sampled utterances (duplicates are omitted for clarity).
}

\end{figure*}

\end{document}